\documentclass[runningheads]{llncs}
\usepackage[T1]{fontenc}
\usepackage{graphicx,verbatim}
\usepackage{booktabs}
\usepackage{tabularx}
\usepackage{array}
 \usepackage{amsmath}
 \usepackage{amssymb}
 \usepackage[table]{xcolor}
\begin{document}
%
\title{MAST-Pro: Dynamic Mixture-of-Experts for Adaptive Segmentation of Pan-Tumors with Knowledge-Driven Prompts}

%
\renewcommand{\thefootnote}{} 
\footnotetext{
    * First author. mengrq@shanghaitech.edu.cn\\
    † Co-first author. ssong25@mgh.harvard.edu\\
    ‡ Corresponding authors. Dinggang.Shen@gmail.com, Guo.Ning@mgh.harvard.edu
}

\author{Runqi Meng\textsuperscript{\rm 1,2,*}, Sifan Song\textsuperscript{\rm 2,†}, Pengfei Jin\textsuperscript{\rm 2}, Yujin Oh\textsuperscript{\rm 2}, Lin Teng\textsuperscript{\rm 1}, Yulin Wang\textsuperscript{\rm 1}, Yiqun Sun\textsuperscript{\rm 1}, Ling Chen\textsuperscript{\rm 2}, Xiang Li\textsuperscript{\rm 2}, Quanzheng Li\textsuperscript{\rm 2}, Ning Guo\textsuperscript{\rm 2,‡}, Dinggang Shen\textsuperscript{\rm 1,3,4,‡}}

\institute{ \textsuperscript{\rm 1}School of Biomedical Engineering \& State Key Laboratory of Advanced Medical Materials and Devices, ShanghaiTech University \\
   \textsuperscript{\rm 2}Center of Advanced Medical Computing and Analysis, Massachusetts General Hospital and Harvard Medical School, Harvard University\\
   \textsuperscript{\rm 3}Shanghai United Imaging Intelligence Co., Ltd.\\
   \textsuperscript{\rm 4}Shanghai Clinical Research and Trial Center}
   
\maketitle              
\begin{abstract}
Accurate tumor segmentation is crucial for cancer diagnosis and treatment. While foundation models have advanced general-purpose segmentation, existing methods still struggle with: (1) limited incorporation of medical priors, (2) imbalance between generic and tumor-specific features, and (3) high computational costs for clinical adaptation. To address these challenges, we propose MAST-Pro (\textbf{M}ixture-of-experts for \textbf{A}daptive \textbf{S}egmentation of pan-\textbf{T}umors with knowledge-driven \textbf{Pro}mpts), a novel framework that integrates dynamic Mixture-of-Experts (D-MoE) and knowledge-driven prompts for pan-tumor segmentation. Specifically, text and anatomical prompts provide domain-specific priors, guiding tumor representation learning, while D-MoE dynamically selects experts to balance generic and tumor-specific feature learning, improving segmentation accuracy across diverse tumor types. To enhance efficiency, we employ Parameter-Efficient Fine-Tuning (PEFT), optimizing MAST-Pro with significantly reduced computational overhead. Experiments on multi-anatomical tumor datasets demonstrate that MAST-Pro outperforms state-of-the-art approaches, achieving up to a 5.20\% improvement in average DSC while reducing trainable parameters by 91.04\%, without compromising accuracy.

\keywords{Pan-tumor Segmentation  \and Foundation model \and Mixture-of-Expert.}

\end{abstract}
\section{Introduction}
Cancer remains a leading cause of mortality worldwide, with incidence and death rates continuing to rise \cite{background}. Since cancer originates from diverse tumor types, early and accurate tumor segmentation is crucial for improving patient outcomes. However, existing methods \cite{NaMa,specifictumor,braintumor} are often task-specific, failing to capture shared tumor characteristics and limiting their scalability in large-scale clinical applications. Therefore, developing a unified pan-tumor segmentation model is essential to enhance diagnostic efficiency and facilitate cross-tumor knowledge transfer. However, there are several challenges in the pan-tumor segmentation task, 
which includes two points: 1) the inherent heterogeneity of tumors across anatomical regions, \textit{i.e.}, exhibiting remarkable diversity of tumors in shape, texture, and intensity, which hinders the adaptability; and 2) the pervasive imbalance in medical datasets, \textit{i.e.}, imbalance distribution of medical datasets for robust feature learning (particularly for rare tumor types), which makes accurate tumor segmentation a daunting task. 


Recently, inspired by foundation models such as the Segment Anything Model (SAM) \cite{SAM} and Contrastive Language-Image Pre-training (CLIP) \cite{CLIP}, prompt-driven approaches \cite{MED_SAM,SegVol,UniversalModel,Zept,CancerUniT} have shown promising performance in medical image segmentation. These methods can be broadly categorized into vision-prompt-driven and text-prompt-driven models. On the one hand, vision-prompt-driven models \cite{MED_SAM,SegVol} leverage visual cues (\textit{e.g.}, points, bounding boxes) to guide segmentation tasks. While effective, these models heavily depend on manual annotations and fail to incorporate anatomical and radiological priors, which are essential for addressing the high heterogeneity of tumors across different anatomical sites. On the other hand, text-prompt-driven methods \cite{UniversalModel,Zept,textual_model} align image and text features within a shared latent space to enhance segmentation across diverse targets. However, their reliance on predefined text templates limits their ability to capture the extensive variability in tumor morphology and radiological presentation, making them less effective in handling domain shifts across anatomical regions. This issue is further compounded by dataset imbalance, resulting in inadequate feature learning and suboptimal segmentation performance, particularly for underrepresented tumor types.

To deal with the imbalance distribution of medical datasets, recent works  \cite{Zept,CancerUniT,CAT} introduced a query-disentangling and self-prompting model to disentangle queries into organ-level and tumor-specific prompts. While this approach represents a step forward, they often overlook shared morphological patterns across tumor types that enhance feature learning for rare tumors \cite{CANCER}, such as edge irregularities and contrast variations. Furthermore, striking a balance between generic and tumor-specific representations still remains challenging, as models struggle to simultaneously retain generic features across anatomical sites while preserving unique tumor characteristics. Beyond accuracy, computational efficiency is another major bottleneck—many approaches either train from scratch on small datasets, failing to leverage large-scale medical imaging data, or rely on full-model fine-tuning \cite{totalsegmentator}, incurring high computational costs and over-fitting risks. Therefore, a scalable and adaptive model is urgently needed to achieve robust and efficient pan-tumor segmentation, addressing both tumor heterogeneity and dataset imbalance while maintaining computational efficiency.

To overcome the aforementioned limitations, we propose MAST-Pro (\textbf{M}ixture-of-experts for \textbf{A}daptive \textbf{S}egmentation of pan-\textbf{T}umors with knowledge-driven \textbf{Pro}mpts), a novel framework that integrates Dynamic Mixture-of-Experts (D-MoE) and knowledge-driven prompts for robust pan-tumor segmentation across diverse anatomical sites. Specifically, to enhance cross-tumor generalization, text and anatomical embeddings derived from higher-order medical knowledge are incorporated as domain-specific priors, guiding the segmentation process. To simultaneously capture generic tumor characteristics and preserve tumor-specific variations, we introduce a dynamic expert selection mechanism, which adaptively allocates computational resources to improve segmentation performance across heterogeneous datasets. Furthermore, we employ Parameter-Efficient Fine-Tuning (PEFT) for multi-anatomical tumor segmentation, significantly reducing computational overhead while enabling efficient adaptation to new tumor types and anatomical regions. Extensive experiments on assembly of eight public datasets demonstrate that MAST-Pro achieves performance comparable to State-of-the-Art (SOTA) methods.

    
    
    


\begin{figure}
\includegraphics[width=\textwidth]{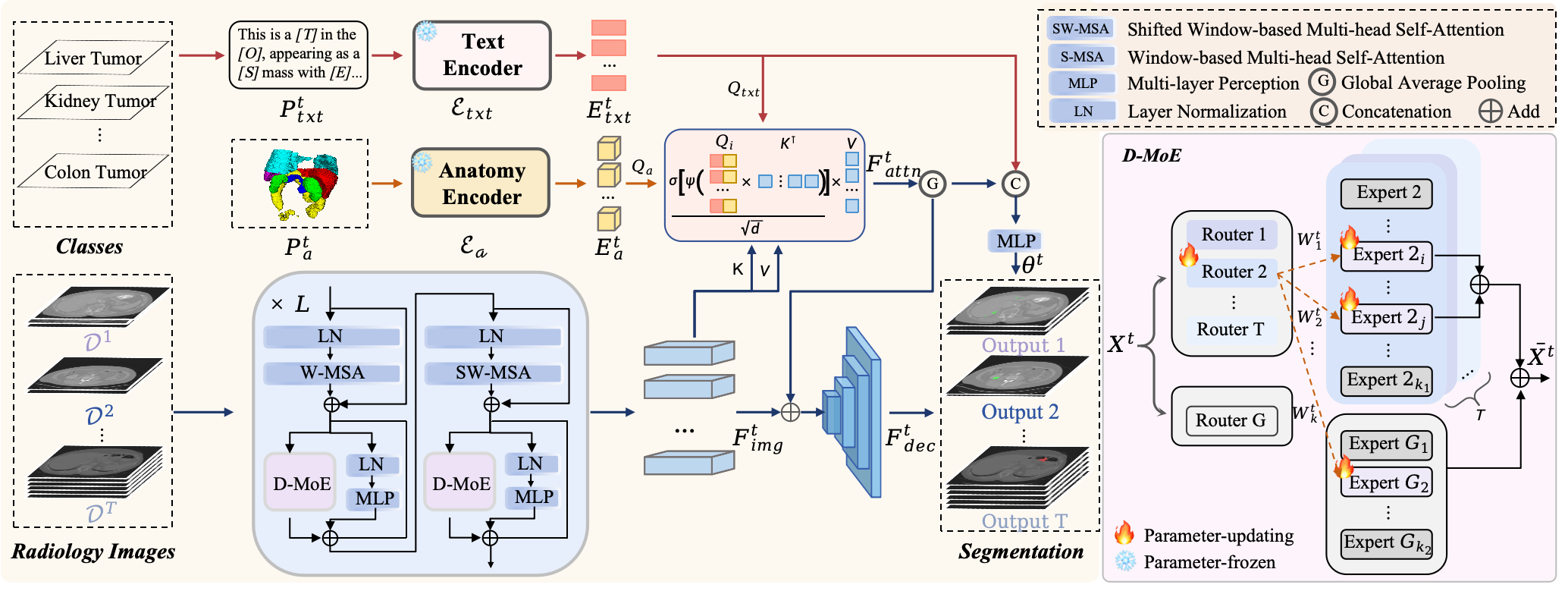}
\caption{An overview of the proposed MAST-Pro model for pan-tumor segmentation, with the text and anatomy prompts served as specific priors to enhance tumor representation learning. Multi-anatomical radiology images are processed through D-MoE-enhanced SwinUNETR, where task-dependent routers dynamically select experts to balance generic and tumor-specific feature learning.} \label{method}
\end{figure}

\section{Method}
In this paper, we propose a novel universal model, called MAST-Pro, for pan-tumor segmentation. As illustrated in Fig. \ref{method}, our approach leverages text ($P_{\text{txt}}^t$) and anatomical ($P_{\text{a}}^t$) prompts as domain-specific priors, incorporating structured medical knowledge to improve segmentation performance (Sect. \ref{sec:subsection1}). Multi-anatomic radiology images are processed through D-MoE-enhanced SwinUNETR, where task-dependent routers dynamically select a mixture of generic and tumor-specific experts to optimize feature learning (Sect. \ref{sec:subsection2}). The extracted prompts are fused with image features, contributing to both segmentation refinement and mask proposal generation. Notably, to improve training efficiency, we first pretrain a backbone on large-scale medical datasets, followed by fine-tuning using PEFT strategy, \textit{i.e.}, up-dating only a small subset of experts within D-MoE rather than the entire model (Sect. \ref{sec:subsection3}).


\subsection{Prompt Embedding}\label{sec:subsection1}
\textbf{Text Prompt Embedding.}  
Each tumor type exhibits distinct characteristics, necessitating a structured text representation enriched with medical domain knowledge. To achieve this, we leverage a large language model (LLM) to generate concise yet informative text descriptions for each tumor type, following a standardized template: $P_{txt}$ = "This is a [C] in the [O], appearing as a [S] mass with [E] borders on [M].", where [C], [O], [S], [E], and [M] represent placeholders for tumor-specific attributes, including tumor type, anatomical location, shape descriptor, edge characteristics, and imaging modality, respectively. These text prompts $\{P_{txt}^t\}$ are then processed using a pre-trained text encoder \cite{CLIP}, denoted as $\mathcal{E}_{txt}$, to extract meaningful feature representations $E_{txt}^t$:

\begin{equation}
E_{txt}^t = \mathcal{E}_{txt}(P_{txt}^t)
\end{equation}


\textbf{Anatomical Prompt Embedding.} Incorporating anatomical priors enhances the model’s ability to recognize tumor characteristics across different anatomy by providing structured spatial information. Given the strong segmentation performance of existing foundation  models in organ segmentation, we leverage organ masks generated by TotalSegmentator \cite{totalsegmentator} as anatomical prompts.The anatomical prompt embedding  $E_{a}$ is obtained by encoding $\{P_{a}^t\}$ using a pre-trained anatomical encoder \cite{swinunetr} denoted as $\mathcal{E}_{a}$:  

\begin{equation}
E_{a}^t = \mathcal{E}_{a}(P_{a}^t)
\end{equation}


\subsection{Pan-tumor Adaptive Mixture-of-Experts}\label{sec:subsection2}
Integrated into the SwinUNETR block, D-MoE enhances pan-tumor segmentation by dynamically selecting experts to balance generic and tumor-specific features. The generalized router $R^g$ directs inputs to shared low-rank experts, capturing foundational morphological and intensity patterns for improved generalization. Meanwhile, the tumor-specific router $R^t$ adaptively selects a combination of generic and tumor-specific experts, optimizing feature representation for each tumor type.

Given a segmentation task $\mathcal{T}^t$, the router adaptively selects the top-$k$ experts based on extracted feature representations:

\begin{equation} \bar{X}^t = \sum_{i=1}^{k} R^{t}_i (X^t) \cdot \left( E^t_i (X^t) ; E^g_i (X^t) \right), \end{equation}
where $X^t$ represents the input feature representation before expert adaptation, and $\bar{X}^t$ is the refined feature representation for task $t$ after processing through D-MoE. The selection weight $R^{t}_i (X^t)$ determines the contribution of each expert. $E^t_i (X^t)$ and $E^g_i (X^t)$ denote the tumor-specific and generic experts, respectively.

The expert selection is determined through a gated mechanism, where the router assigns selection weights via:

\begin{equation} R(X) = \text{Softmax}(\text{KeepTop-}k(X^\top W, k)), \end{equation}
where $W$ is a learnable projection matrix, and KeepTop-$k$ retains the highest-weighted expert activations:

\begin{equation}
\text{KeepTop-}k(v, k)_i =
\begin{cases}
    v_i, & \text{if } v_i \text{ is in top } k \text{ elements of } v, \\
    -\infty, & \text{otherwise}.
\end{cases}
\end{equation}

To integrate domain-specific priors into visual representations, the extracted prompts act as queries $\mathcal{Q}_j$, while the image features serve as keys $\mathcal{K}$ and values $\mathcal{V}$ within the cross attention mechanism, formulated as:
\begin{equation}
\mathcal{F}_{attn} = \text{Softmax} \left( \frac{[\mathcal{Q}_{txt}; \mathcal{Q}_{a}]  \mathcal{K}^\top}{\sqrt{d}} \right) \mathcal{V}, 
\end{equation}
where $d$ is a scaling factor for stability. The attention-refined features ($\mathcal{F}_{attn}$) are fused with image features ($\mathcal{F}_{img}$) and decoded by SwinUNETR:

\begin{equation} \mathcal{F}_{dec} = \mathcal{D} (\mathcal{F}_{attn} + \mathcal{F}_{img}). \end{equation}

To further refine contextual understanding, $\mathcal{F}_{attn}$ undergoes global average pooling (GAP) and is concatenated with text prompts. The resulting feature vector is then processed by a multi-layer perceptron (MLP) to generate an initial mask proposal $\theta^t$, supervised via Cross-Entropy loss with tumor category information. Finally, the mask proposal $\theta^t$ then guides the final output ($\mathcal{F}_{final}$), while Dice Loss further refines tumor segmentation.

\begin{equation}
\mathcal{F}_{final} = \mathcal{F}_{dec} \theta^t, \quad \text{where} \quad
\theta^t = \text{MLP}(\text{GAP}(\mathcal{F}_{attn}); \mathcal{E}_{txt}).
\end{equation}

\subsection{Parameter-Efficient Fine-Tuning (PEFT) strategy}\label{sec:subsection3}
In D-MoE, task-dependent routers selectively fine-tune low-rank experts to preserve generic tumor characteristics while refining tumor-specific features. Specifically, we employ $k_2$ experts to encode generic tumor features and $k_1$ experts to capture domain-specific variations. The routers dynamically select the top $k$ experts from both groups, enabling adaptive feature learning.

\section{Experiment}
\subsection{Dataset}

Following prior work \cite{UniversalModel}, we pretrained our model's backbone on a diverse collection of large-scale medical imaging datasets, including BTCV \cite{BTCV}, CT-ORG \cite{ctorg}, Pancreas-CT \cite{Pancreas-CT}, CHAOS \cite{CHAOS}, 3D-IRCADb \cite{3D-IRCADb}, WORD \cite{WORD}, and AMOS \cite{AMOS}, along with tumor-specific datasets such as AbdomenCT-1K \cite{Abdomenct-1k}, LiTS \cite{LiTS}, KiTS \cite{KiTS}, and CT images from the MSD dataset \cite{MSD}. Building on this foundation, we curated 2,000+ tumor cases from eight tumor-specific datasets to develop a robust pan-tumor segmentation model.

To ensure consistency, all CT scans were reoriented, resampled to $1.5 \times 1.5 \times 1.5 mm^3$ isotropic spacing, and cropped to focus on tumor-relevant regions. During training, we extracted $96 \times 96 \times 96$ voxel patches, ensuring balanced tumor and background sampling. Data augmentation included random 90-degree rotations and intensity shifting to enhance robustness and generalization.


\subsection{Implementation Details}
All experiments were conducted using PyTorch on four NVIDIA Tesla H100 GPUs (80GB RAM). The backbone was pretrained for 1,000 epochs, followed by PEFT applied to D-MoE. Both pretraining and fine-tuning were performed with identical hyperparameters, utilizing the AdamW optimizer with a base learning rate of $5 \times 10^{-5}$ and a batch size of 4.  Multi-GPU training was performed using Distributed Data Parallel (DDP) to ensure efficient scalability.

\subsection{Comparison with State-of-the-art Methods}

To evaluate the effectiveness of our method, we conducted comparative experiments against SOTA universal medical image segmentation approaches, which are categorized into baseline methods, vision-prompt-based methods, and automatic text-prompt-based methods. Specifically, we selected nnUNet (3D full resolution) \cite{nnunet} and Swin UNETR \cite{swinunetr} as baselines, Med-SAM3D \cite{MED_SAM}, MA-SAM \cite{masam}, and SegVol \cite{SegVol} as vision-prompt-based methods, and the Universal Model \cite{UniversalModel} and ZePT \cite{Zept} as automatic text-prompt-based models.

\textbf{Quantitative Comparison.}  Table \ref{res1} presents the segmentation performance across eight datasets, demonstrating that MAST-Pro achieves the highest mean DSC of 68.71\%, outperforming both vision-prompt-based (\textit{e.g.}, Med-SAM3D, SegVol) and text-prompt-based (\textit{e.g.}, ZePT, Universal Model) approaches. Compared to the strongest baseline, ZePT, our model improves the average DSC by 5.2\%, highlighting the effectiveness of D-MoE and knowledge-driven prompts in enhancing generalization. Our model achieves top performance on six out of eight datasets, with notable improvements in M-Pa (+4.26\%), M-HT (+4.11\%), and LiTS (+2.46\%), demonstrating its robustness across diverse tumor types. Particularly in Liver tumor segmentation (LiTS), MAST-Pro surpasses Med-SAM3D by 59.8\% and SegVol by 19.94\%, showcasing its ability to capture tumor-specific features autonomously. 


\begin{table}[t]
\centering
\fontsize{8}{11}\selectfont
\caption{DSC (\%) results for multimodal segmentation methods across all datasets. Comparison of segmentation performance on internal and external datasets. The best results in each column are bolded. Abbreviations: "M-Li" – "MSD-Liver", "M-Lu" – "MSD-Lung", "M-Pa" – "MSD-Pancreas", "M-HT" – "MSD-HepaticVessel Tumor", "M-Co" – "MSD-Colon", "Abd" – "AbdomenCT-1K".}
\setlength{\tabcolsep}{1mm} 
\label{res1}
{
\begin{tabular}{l | c c c c c c c c |c }
\hline
\textbf{Method} & \textbf{M-Li} & \textbf{M-Lu} & \textbf{M-Pa} & \textbf{M-HT} & \textbf{M-Co} & \textbf{LiTS} & \textbf{KiTS} & \textbf{Abd} & \textbf{Mean} \\
\hline
nnU-Net \cite{nnunet} & 60.22 &68.54  & 52.75 & 69.50 & 45.07 & 57.15 & 65.18 & 62.85 & 60.16 \\
Swin UNETR \cite{swinunetr} & 63.24 & 66.70 & 53.24 & 66.23 & 42.55 & 66.79 & 65.23 & 64.82 & 61.10  \\
\hline
Med-SAM3D \cite{MED_SAM} & 47.81 & 24.28 & 40.26 & 57.89& 48.21 & 22.32 & 67.15 & - & 43.98  \\
MA-SAM \cite{masam} & 69.16 & 51.70  & 31.22 & 63.57 & 39.98 & 57.22 & \textbf{75.91} & - & 55.53 \\
SegVol \cite{SegVol} & 69.07 & 65.53 & 54.35 & 68.75 & \textbf{48.23} & 62.18 & 57.74 & - & 60.84 \\
\hline
Universal Model \cite{UniversalModel} & 65.92 & 67.11  & 54.72 & 66.31 & 42.82 & 76.07 & 62.86 & 66.53  & 62.79 \\
ZePT \cite{Zept} & 69.58 & 69.07 & 53.39 & 70.65 & 43.18 & 79.66 & 57.83 & 64.76 & 63.51\\

\textbf{Ours} & \textbf{72.96} & \textbf{72.10}  & \textbf{59.34} & \textbf{74.76} & 46.79 & \textbf{82.12} & 72.99 & \textbf{68.65} & \textbf{68.71} \\
\hline
\end{tabular}
}
\end{table}

\textbf{Qualitative Comparison. } Fig. \ref{res2} presents qualitative segmentation results, demonstrating the superiority of our method in capturing tumor boundaries and preserving structural details. To be specific, our method demonstrates superior segmentation accuracy, capturing finer tumor details with fewer false positives and better boundary adherence compared to the competing methods. Particularly in small or complex tumors (first and second rows), MAST-Pro closely matches the ground truth, whereas other models either under-segment (ZePT, Universal Model) or over-segment (MA-SAM, SegVol) tumors, leading to inaccuracy. Additionally, in large and irregular tumors (third and fourth rows) our model produces more accurate contours and reduces misclassification errors, highlighting its robustness in handling diverse tumor morphology.


\begin{figure}
\centering
\includegraphics[width=0.9\textwidth]{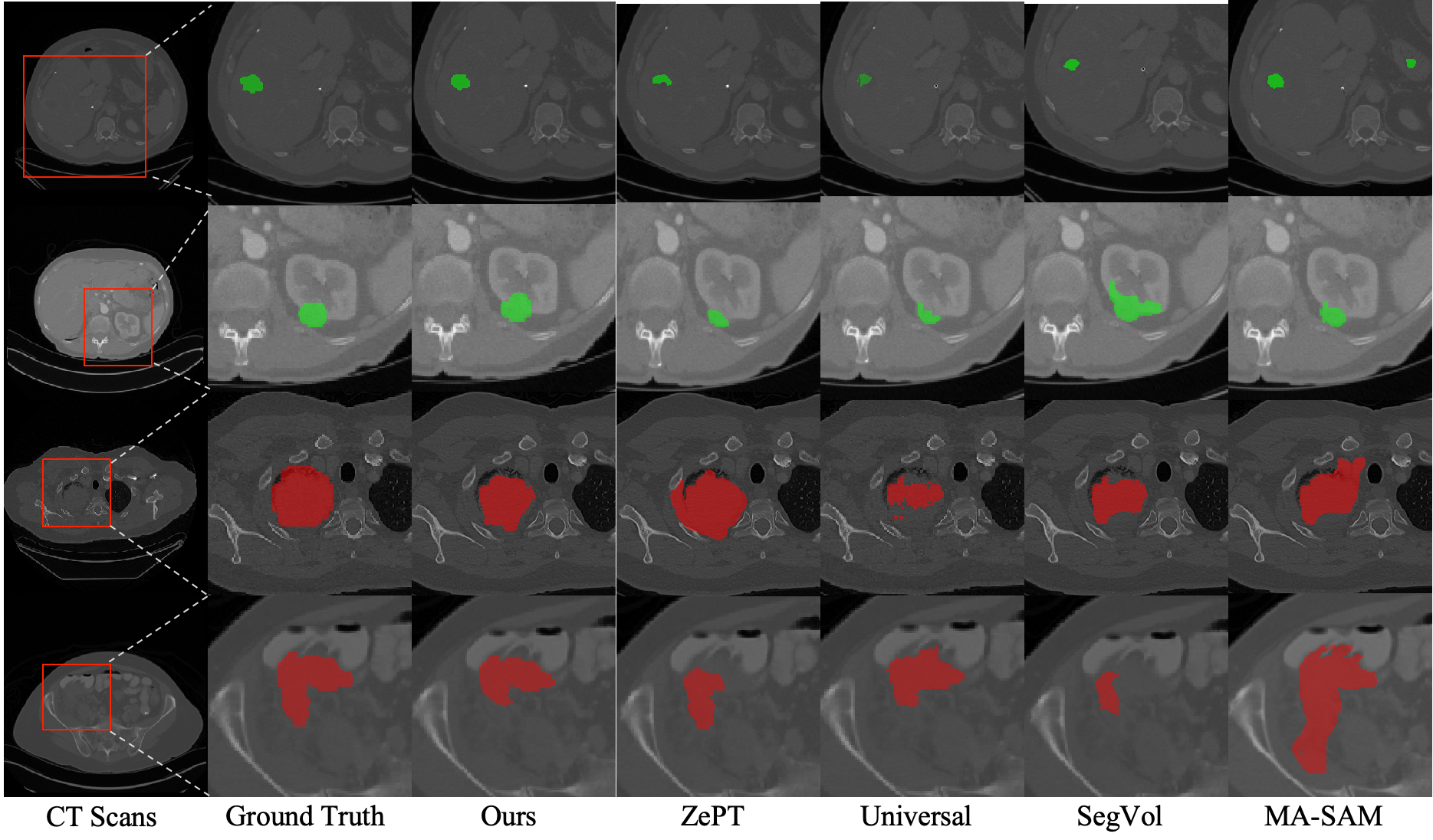}
\caption{Qualitative visualizations comparing the proposed MAST-Pro model with other prompt-driven methods for multi-tumor segmentation. The first column shows the original CT scans, while the second column presents the ground-truth segmentations. The segmentation results in rows one to four correspond to liver tumors, kidney tumors, lung tumors, and colon tumors, respectively.} \label{fig1}
\end{figure}

\subsection{Ablation Study}

\begin{table}[t]
\centering
\fontsize{8}{11}\selectfont
\caption{Ablation study on the effectiveness of anatomical prompts (AP), text prompts (TP), and the D-MoE module. The results are reported in DSC (\%). "\checkmark" indicates the inclusion of a component, while "$\times$" denotes its absence.}
\setlength{\tabcolsep}{1mm} 
\label{res2}
{
\begin{tabular}{ c c c| c c c c c c c c |c }
\hline
\textbf{AP} & \textbf{TP} &\textbf{D-MoE} &\textbf{M-Li} & \textbf{M-Lu} & \textbf{M-Pa} & \textbf{M-HT} & \textbf{M-Co} & \textbf{LiTS} & \textbf{KiTS} & \textbf{Abd} & \textbf{Mean} \\

$\times$ & $\times$ & $\times$  & 63.24 & 66.70 & 53.24 & 66.23 & 42.55 & 66.79 & 65.23 & 64.82 & 61.10  \\
\checkmark & $\times$ & $\times$  & 63.45 & 70.46  & 56.82 & 68.92 & 44.82 & 72.72 & 68.44& 62.16 & 63.47\\

 $\times$ & \checkmark & $\times$ & 66.48 & 68.24  & 53.42 & 68.02 & 42.26 & 77.03 & 64.86 & 67.08  & 63.42 \\
$\times$ & $\times$& \checkmark & 67.45 & 69.21 & 56.20 & 66.24 & 44.25 & 72.16 & 63.42 & 65.13 & 63.01  \\
\checkmark & \checkmark & \checkmark & \textbf{72.96} & \textbf{72.10}  & \textbf{59.34} & \textbf{74.76} & \textbf{46.79} & \textbf{82.12} & \textbf{72.99} & \textbf{68.65} & \textbf{68.71}   \\
\hline
\end{tabular}
}
\end{table}

Table \ref{res2} shows the ablation study, evaluating the contribution of each component.

\textbf{Effect of Knowledge-Driven Prompts.} Removing all prompts leads to a significant drop in performance (61.10\% mean DSC), underscoring their importance.  Anatomical prompts alone improves segmentation (+2.27\% mean DSC), particularly in M-Li and M-Pa, by incorporating structural priors. Text prompts alone enhances performance in LiTS but struggles with fine-grained boundaries, highlighting its limitations in handling tumor variability.

\textbf{Effect of D-MoE.} Introducing D-MoE alone improves the mean DSC to 63.01\%, particularly benefiting LiTS (+5.37\%) and M-Li (+4.21\%), demonstrating its ability to balance generic and tumor-specific features. However, without domain priors, its effectiveness is constrained in highly variable datasets.

\textbf{Computational Efficiency.}
To assess computational efficiency, we compare trainable parameters and GPU memory usage during training. As shown in Table \ref{tab:computational_cost}, our model requires only 21.04M parameters and the lowest GPU memory, demonstrating the effectiveness of PEFT in reducing computational overhead while maintaining high segmentation accuracy.

\begin{table}[h]
\centering
\caption{Comparison of computational cost during training between our method and others in terms of training parameters and GPU memory usage.}
\label{tab:computational_cost}
\fontsize{8}{11}\selectfont
\setlength{\tabcolsep}{1mm} 
\begin{tabular}{lccc}
\toprule
\textbf{Method} & \textbf{Train Params} $\downarrow$  & \textbf{Memory Usage}  $\downarrow$ \\
\midrule

MA-SAM & 363.68 M& 74544.02 MB \\
SegVol & 449.08 M & 19898.00 MB  \\
Universal Model & 244.80 M& 9710.55 MB \\
ZePT & 495.10 M & 24705.40 MB\\
Ours & 21.04 M  & 8961.30 MB& \\

\bottomrule
\end{tabular}
\end{table}

\section{Conclusion}
In this paper, we propose MAST-Pro, a novel framework that integrates Dynamic Mixture-of-Experts (D-MoE) and knowledge-driven prompts for pan-tumor segmentation. Text and anatomical prompts provide domain-specific priors, while D-MoE dynamically balances generic and tumor-specific feature learning, improving segmentation across diverse tumor types. Additionally, Parameter-Efficient Fine-Tuning (PEFT) reduces computational overhead without compromising accuracy. Experiments on multi-anatomical tumor datasets show MAST-Pro outperforms state-of-the-art methods by 5.20\% DSC while reducing trainable parameters by 91.04\%, demonstrating its effectiveness in accurate, generalizable, and efficient tumor segmentation.

%
%
%
%

\end{document}